\documentclass[11pt,a4paper]{article}
\usepackage[hyperref]{naaclhlt2019}
\usepackage{times}
\usepackage{latexsym}
\usepackage{amsmath}
\usepackage{amssymb}
\usepackage{url}
\usepackage{multirow}
\usepackage{adjustbox}
\usepackage{algorithm}
\usepackage[noend]{algpseudocode}
\usepackage{pgfplots}

\aclfinalcopy 

\usepackage{caption}
\usepackage{booktabs}

\title{Learning Invariant Representations for Sentiment Analysis: \\ The Missing Material is Datasets}

\author{Victor Bouvier \\
  Centrale Sup\'elec \\ Universit\'e Paris-Saclay \\
  Sidetrade \\
  {\tt \small vbouvier@sidetrade}\And
  C\'eline Hudelot \\
  Centrale Sup\'elec \\ Universit\'e Paris-Saclay \\
  {\tt \small celine.hudelot@centralesupelec.fr} \\ \AND
    Philippe Very\\
  Sidetrade  \\
  {\tt \small pvery@sidetrade.com} \\ \And
Cl\'ement Chastagnol\\
  Sidetrade\\
  {\tt \small cchastagnol@sidetrade.com}  
}

\begin{document}
\maketitle
\begin{abstract}
Learning representations which remain invariant to a nuisance factor has 
a great interest in Domain Adaptation, Transfer Learning, and Fair 
Machine Learning. Finding such representations becomes highly challenging in NLP tasks since the nuisance factor is entangled in a raw text. To our knowledge, a major issue is also that only few NLP datasets allow assessing the impact of such factor. In this paper, we introduce 
two generalization metrics to assess model robustness to a  nuisance 
factor: \textit{generalization under target bias} and 
\textit{generalization onto unknown}. We combine those metrics with a  simple data filtering approach to control the impact of the nuisance factor on the data and thus to build experimental biased datasets. We 
apply our method to standard datasets of the literature (\textit{Amazon} 
and \textit{Yelp}). Our work shows that a simple text classification baseline (i.e., sentiment analysis on reviews) may be badly affected by the \textit{product ID} (considered as a nuisance factor) when learning the polarity of a review. The method proposed is generic and applicable as soon as the nuisance variable is annotated in the dataset.
\end{abstract}

\section{Introduction}

Significant improvements have been made recently on various Computer Vision \cite{krizhevsky2012imagenet} and Natural Language Processing \cite{glorot2011domain} tasks using deep architectures. These successes are mainly explained by the capacity of specific neural architectures to learn internal representations which  remain insensitive to transformations such as translations and rotations of an image or sequencing of words for a text \cite{bengio2006curse,bengio2009learning}. However, remaining invariant to change in latent factors (called nuisance factor) is more difficult and requires an additional modeling effort. Current approaches consist in considering the nuisance factor as a random variable and constraining during learning the model to be invariant to this variable. For instance, in a context of Domain Adaptation \cite{quionero2009dataset, pan2010survey}, \cite{ganin2014unsupervised} suggest to learn an invariant representation to source and target domains by modeling the domain as a binary nuisance variable. Also, Learning Fair Representations \cite{zemel2013learning} aims to learn representation invariant to an unwanted factor (e.g. gender for the \textit{Adult} dataset and age for the \textit{German} dataset \cite{Dua:2017}).

There is no consensus on how to measure invariance of learned representation to  nuisance factors. Both Domain Adaptation and Fair Representation communities suggest to evaluate the amount of information on the nuisance factor held by the representation (\textit{identifiability}). It consists in training a classifier to map the learned representation to the nuisance factor \cite{ganin2014unsupervised, zemel2013learning, xie2017controllable} and in reporting its accuracy. In order to assess invariance of predictions, the \textit{discrimination} metric \cite{kamiran2009classifying, kamishima2011fairness} consists in identifying for each label a protected group (positively biased to the label) and an unprotected group (negatively biased to the label) and computing the shift in label ratio prediction for the two groups.

Moreover, very few datasets allow to measure the impact of nuisance factors on machine learning models.  To our knowledge, the dataset mentioned in \cite{blitzer2011domain} is a rare exception in the sentiment analysis community. This dataset allows to assess the impact of four topics on reviews. It provides however only few samples (1,000 labeled to 2,000 unlabeled) by topic.

In the present work, we show that both \textit{identifiability} and \textit{discrimination} may not reflect the model robustness to a nuisance factor. For this reason, we introduce two additional metrics. \textit{Generalization under target bias} consists in training a model on a train set where the label and the nuisance factor are dependent and reporting its accuracy on a test set where the nuisance factor and the label are independent. \textit{Generalization onto unknown} assesses the model accuracy on a test set where the nuisance factor takes values that were absent in the train set.

Besides, to tackle the issue of the lack of adequate datasets, we suggest a simple data filtering approach applied to existing datasets for evaluating these metrics. In our filtering method, we build a sequence of subsets which exhibit an increasing dependency between the nuisance factor and features while keeping the label and the nuisance factor independent. We apply our method for evaluating the robustness of a popular text classification algorithm \cite{joulin2016bag}. We used four datasets \textit{Amazon} (Book, Electronics, Musics) and \textit{Yelp} for which we considered \textit{product ID} as a nuisance factor for learning polarity of a review.

\section{Assessing robustness to nuisance factor}
\subsection{Formalization}
\paragraph{Notations}
Let us denote $V$ a random variable, its realization $v$ and its domain $\mathcal V$.  For a couple of random variables $(U,V)$ with probability $\mathbb  P(U,V)$, we note $ I_{\mathbb P}(U,V) = \mathbb E_U[- \log d\mathbb P(U)] - \mathbb E_{U,V} [ - \log d \mathbb P(U|V)]$ the mutual information between $U$ and $V$. We consider the case of a dataset $\mathcal{D}$ which is a set of realizations of a random triplet $(X,Y,S)$ with probability $\mathbb P(X,Y,S)$ where $X$ is the feature, $Y$ the label and $S$ the nuisance factor.

\paragraph{Learning an invariant model}  
 The Invariant Representation Learning problem consists in learning a representation $H$ of $X$ such that the mapping from $H$ to $Y$ is invariant to $S$. Difficulties emerge when $X$ and $Y$ are dependent from $S$. In this work, dependency is measured with mutual information.

\subsection{Existing measures of sensibility to a nuisance factor}
\paragraph{Identifiability} After having trained a model which maps $X$ to $Y$ through an internal representation $H$, \textit{identifiability} metric suggests to evaluate the amount of information in $H$ about $S$. It consists in training a second model which maps $H$ to $S$ and reporting its accuracy \cite{ganin2014unsupervised, zemel2013learning}.

\paragraph{Discrimination} After having trained a model $\hat{\mathbb P}$ which maps $X$ to $Y$, the \textit{discrimination} \cite{kamiran2009classifying, kamishima2011fairness, zemel2013learning} metric works on two predetermined \textit{protected} and \textit{unprotected} groups. For a label $y \in \mathcal Y$, the protected group is defined by $\mathcal D_{\mathrm{pr}(y)} = \{ (x,y,s) \in \mathcal D \backslash \arg \max_{y' \in \mathcal Y} \mathbb P(y'|s) = y \} $
and the unprotected group is defined as $\mathcal D_{\mathrm{upr}(y)} = \mathcal D \backslash \mathcal D_{\mathrm{pr}(y)}$. Then $
    \mathrm{disc}(y) = \left| \frac{|\hat {\mathcal D}_\mathrm{pr}(y)| }{|\mathcal D_{\mathrm{pr}(y)}|}  \right. 
    - \left. \frac{|\hat{\mathcal D}_\mathrm{upr}(y)| }{|\mathcal D_{\mathrm{upr}(y)}|}\right | $ 
where $\hat {\mathcal D}_\mathrm{pr}(y) = \{(x,y,s) \in \mathcal D_\mathrm{pr} \backslash y = \hat y (x) \}$, $\hat {\mathcal D}_\mathrm{upr}(y) = \{(x,y,s) \in \mathcal D_\mathrm{upr} \backslash y = \hat y (x) \}$ and $\hat y(x) = \arg \max_{y' \in \mathcal Y} \hat{\mathbb P}(y'|x)$. It quantifies how a model tends to predict label $y$ on the protected class (which is favorable to $y$) compared with the unprotected class (which is unfavorable to $y$).

\subsection{Proposal of two additional metrics}
Rather than computing a metric on a test set following the same distribution as the train set, we suggest to report the accuracy of the model on a well-designed test set. Such train / test split depends on the desired model behavior we want to track. We then introduce the two following metrics.

\paragraph{Generalization under target bias (GTB)} We propose to evaluate model robustness to target bias to the nuisance variable $S$. For instance suppose that in a loan granting classification problem, loans are effectively granted more easily to men in the train set. We want to assess how a trained model behaves on a test set where men and women are treated equally. More formally, such bias is a situation where the training data $(X,Y)$ is drawn from $
    \mathbb P(X,Y) = \mathbb E_{S \sim \mathbb P(S)} [\mathbb P(Y|S) \mathbb P(X|Y,S) ]$ and testing data is drawn from $\mathbb P^\star(X,Y)$ such that $\mathbb P^\star(Y|S) \neq \mathbb P(Y|S)$ while conserving $\mathbb P(S)$ and $\mathbb P(X|Y,S)$. We set $\mathbb P^\star(Y|S)= \mathbb P(Y)$ which is the scenario where $Y$ and $S$ are dependent on the train set but independent on the test set. We suggest to restrict evaluation on family of distributions $(\mathbb P_{\beta, y_s})_{\beta \in [0.5, 1], s \in \mathcal S}$ where $\mathbb P_{\beta}(Y|S) = \beta$ if $Y =y_s$ and $\frac{1 - \beta}{ |\mathcal Y| - 1}$ otherwise. The sampling procedure is detailed in Appendix in Algorithm \ref{alg:target_bias}.

\paragraph{Generalization onto unknown (GU)} We propose to assess the amount of generic knowledge learned by the model by evaluating it on a test set with values of $S$ that were absent during training. For instance, for a movie reviews corpus, the test set is composed of movies unseen during training. More formally, it consists in training on data drawn from $ \mathbb P(X,Y) = \mathbb E_{S \sim \mathbb P(S)} [\mathbb P(Y,X|S) ]
$ and testing data is drawn from $
    \mathbb P^\star(X,Y) = \mathbb E_{S \sim \mathbb P^\star(S)} [\mathbb P(Y,X|S) ]$
where $\mathcal S = \mathcal S' \cup \mathcal S^\star$ with $\mathcal S^\star \cap \mathcal S' = \emptyset$ and $\mathbb P(S) = \frac{1}{|\mathcal S'|}$ if $S \in \mathcal S'$ and $0$ otherwise while $\mathbb P^\star(S) = \frac{1}{|\mathcal S^\star|}$ if $S \in \mathcal S^\star$ and $0$ otherwise. In order to not interfere with \textit{GTB} metric, we suggest to conserve $\mathbb P(Y|S) = \mathbb P^\star(Y|S) = \mathbb P(Y)$ which is the situation where $S$ and $Y$ are independent in both train and test sets.

In order to have a fair comparison between two train / test splits, metrics are reported as the percentage of the accuracy computed on a test set which follows the same distribution than the train set. 

\subsection{Our proposed data filtering approach}
\label{data_filtering}
In order to increase the variety of datasets where metrics are evaluated, we suggest a simple data filtering approach which enables us to increase the dependence of the features $X$ with the nuisance factor $S$ while keeping independence between label $Y$ and the nuisance factor $S$. We suggest to iteratively reject samples of the initial dataset in order to build a sequence of included subsets. The sequence is built such that the dependence between features $X$ and the nuisance factor $S$ increases. It consists in estimating the contribution of each sample to the mutual information between the features and the nuisance factor. For a sample $(x,y,s)$, the contribution $i_{\mathbb P}(x)$ to $I_{\mathbb P}(X,S)$  is defined as $\mathbb E_S[-\log d\mathbb P(S)] - \mathbb E_{S|X}[-\log d\mathbb P(S|X=x)]]$. It is computed by training a mutual information estimator on $\mathcal D \backslash \{(x,y,s)\}$. To reduce computation time, we proceed with $N_b$ batches of data $(\mathcal D_i)_{i=1}^{N_{b}}$  where contribution to mutual information of samples of $\mathcal D_i$ is computed by training an estimator on $\cup_{i \neq n} \mathcal D_i$. We reject an $\alpha-$quantile of the distribution of contributions. For a given nuisance factor and label, the $\alpha-$quantile may not be totally rejected to ensure that $Y$ and $S$ remain independent. The procedure is described in Appendix \ref{sec:data_filtering}.

\section{Experiments}

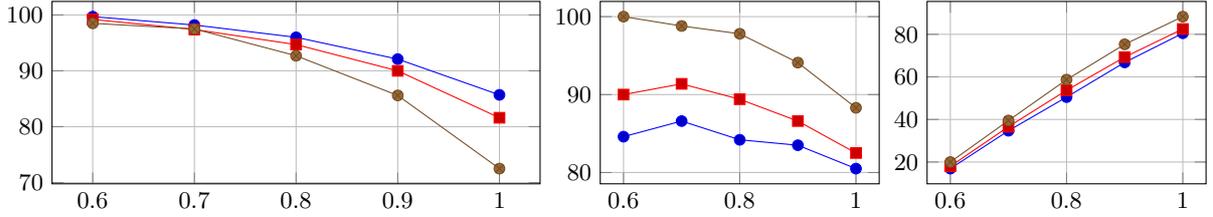
\begin{figure*}
\begin{center}
\footnotesize
\begin{tikzpicture}
	\begin{axis}[
		height=4.cm,
		width=8cm,
		grid=major,
        legend pos=south west
	]
	\addplot coordinates {
	    (0.6, 99.7)
		(0.7, 98.2)
        (0.8, 96)
        (0.9, 92.1)
        (1.0, 85.7)};
	 \label{0}
	
	\addplot coordinates {
	    (0.6, 99.2)
		(0.7, 97.4)
        (0.8, 94.7)
        (0.9, 90)
        (1.0, 81.6)};
    \label{1}
	
	\addplot coordinates {
	    (0.6, 98.5)
		(0.7, 97.5)
        (0.8, 92.7)
        (0.9, 85.6)
        (1.0, 72.5)};
	\label{2}
	
	\label{3}
	\end{axis}
\end{tikzpicture}
\begin{tikzpicture}
	\begin{axis}[
		height=4.cm,
		width=5.25cm,
		grid=major,
        legend pos=south west
	]
	\addplot coordinates {
	    (0.6, 84.6)
		(0.7, 86.6)
        (0.8, 84.2)
        (0.9, 83.5)
        (1.0, 80.5)};
	 \label{0}
	
	\addplot coordinates {
	    (0.6, 90.0)
		(0.7, 91.375)
        (0.8, 89.4)
        (0.9, 86.6)
        (1.0, 82.5)};
    \label{1}
	
	\addplot coordinates {
	    (0.6, 100)
		(0.7, 98.8)
        (0.8, 97.8)
        (0.9, 94.1)
        (1.0, 88.3)};
	\label{2}

	
	\end{axis}
\end{tikzpicture}
\begin{tikzpicture}
	\begin{axis}[
		height=4.cm,
		width=5.25cm,
		grid=major,
        legend pos=south west
	]
	\addplot coordinates {
	    (0.6, 16.9)
		(0.7, 34.7)
        (0.8, 50.5)
        (0.9, 66.8)
        (1.0, 80.5)};
	 \label{0}
	
	\addplot coordinates {
	    (0.6, 18.0)
		(0.7, 36.6)
        (0.8, 53.7)
        (0.9, 69.3)
        (1.0, 82.5)};
    \label{1}
	
	\addplot coordinates {
	    (0.6, 20.0)
		(0.7, 39.5)
        (0.8, 58.7)
        (0.9, 75.3)
        (1.0, 88.3)};
	\label{2}

	\end{axis}
\end{tikzpicture}
\end{center}
\caption{From left to right: \textit{Generalization under target bias}, $\mathrm{disc}(y)/(2\beta-1)$ and $\mathrm{disc(1)}$ according to $\beta$ (iterations: \{$\mathcal D_0$: \ref{0}, $\mathcal D_1$: \ref{1}, $\mathcal D_2$: \ref{2}\}). }
\label{fig:target_bias}
\end{figure*}

\subsection{Baselines}

\paragraph{Comparing metric}
For \textit{GTB} metric and a given $\beta$, additionally to \textit{discrimination} and \textit{identifiability}, we report $\frac{\mathrm{disc}(y)}{|2 \beta -1|}$. This ratio reflects how the model differs from a perfect accuracy-wise model which is invariant to the nuisance factor. Such a model has a \textit{discrimination} of $2\beta-1$.

\textit{Discrimination} is not defined in the context of \textit{GU} metric since $Y$ and $S$ are independent in both train set and test sets which implies there is no protected group. Besides, for \textit{identifiability}, we report accuracy on a classifier trained to separate $\mathcal S'$ to $\mathcal S^\star$ which is the current approach in Domain Adaptation.

\paragraph{Sampling baseline} In order to evaluate the impact of our sampling procedure, we consider the initial dataset as the baseline (i.e. no filtering).

\subsection{Datasets and model}

We used for our experimentations four datasets:  three from \textit{Amazon} \cite{mcauley2013hidden} (Books (AB), Electrics (AE) and Musics (AM)) and \textit{Yelp}\footnote{\url{www.yelp.com/dataset/challenge}}. Those datasets are reviews with an annotated polarity (rating among 1 to 3 are considered as negative while reviews among 4 and 5 are considered as positive). For each review, the \textit{product ID} is annotated. In our experiment, we consider the \textit{product ID} as a nuisance factor for learning the polarity of a review. For instance, one can be interested in learning a model which does not learn from \textit{Amazon} users that \textit{Star Wars} movies are globally recognized as good movies. We suggest to use a selected sample of those datasets considering only 50 \textit{product ID} per datasets (those with the larger number of reviews). Then we sample it again to have the same number of negative and positive polarities per value of the nuisance factor. We chose to assess the robustness a popular text classification model \cite{joulin2016bag} as a working example. Details on hyper-parameters are given in Appendix \ref{sec:training_details}. The representation of a given text is then the mean value of words embedding.

\subsection{Results}
We performed data filtering on 2 iterations with $\alpha = (1/4, 1/3)$ to obtain $\mathcal D_0 \supset \mathcal D_1 \supset \mathcal D_2$. Mutual information was estimated using 10 batches as described in \ref{data_filtering}. We report in Table \ref{datasets_statistics} the number of samples per subsets ($\mathcal D_0$ is the initial dataset) with associated mutual information. By rejecting samples, the mutual information of subsets increases.

\begin{table}[h!]
\centering 
\begin{tabular}{|c|cccc|}
\hline
Dataset & Yelp &  AB  & AE & AM   \\
\hline
\multirow{2}{*}{$\mathcal D_0$} & 103k & 142k & 129k & 101k \\
 & 1.68 & 1.22  & 2.62 & 2.20 \\ 
 \hline 
\multirow{2}{*}{$\mathcal D_1$} & 76k & 105k & 100k & 73k \\
& 2.21 & 2.22 & 3.09 & 2.89  \\ \hline
\multirow{2}{*}{$\mathcal D_2$} & 48k & 68k & 71k & 48k   \\
& 2.86 & 2.60 & 3.34 & 3.15 \\
\hline
\end{tabular}

\caption{Number of samples and mutual information estimation for each subsets and its associated mutual information.  }
\label{datasets_statistics}
\end{table}

We set $\beta \in \{0.6, 0.7, 0.8, 0.9, 1.0\}$ and we report aggregated \textit{GTB} metric and \textit{discrimination} taking the mean on the four datasets. \textit{Discrimination} increases mechanically with both the mutual information and $\beta$ making it hard to interpret. $\mathrm{disc}(y)/(2 \beta -1)$ allows to observe a clear separation: this ratio increases with the mutual information but decreases with $\beta$. This inverted monotony is counter-intuitive: one may expect the metric to decrease as mutual information and $\beta$ increase since the problem seems harder to address. \textit{GTB} metric follows this intuition. We did not report any significant result for \textit{identifiability} metric: the accuracy to identify the nuisance factor was lower than 2\% more than the majority class.

For \textit{GU} metric, we report the mean value accuracy and standard deviation for the $50$ different random splits of  $\mathcal S = \mathcal S' \cup \mathcal S^\star$ with $|\mathcal S' | = |\mathcal S^\star|= 25$. Mean and standard deviation of \textit{Identifiability} is computed for 10 random splits. \textit{GU} decreases while mutual information increases (see Table \ref{tab:unknown}). It appears the text classification model studied is robust to new values of the nuisance factor.
\begin{table}[h!]
    \centering
    \begin{tabular}{|c|ccc|}
    \hline
        Dataset & $\mathcal D_0$ & $\mathcal D_1$ & $\mathcal D_2$  \\
    \hline
       \multirow{2}{*}{Yelp} & 99.4$\pm$0.5 & 99.0$\pm$1.0  & 99.0$\pm$1.0        \\
       & 56.5$\pm$3.4 & 55.6$\pm$5.0  & 57.6$\pm$4.0        \\ \hline

        \multirow{2}{*}{AB} & 98.3$\pm$0.7 & 97.6$\pm$0.9&   97.2$\pm$1.2      \\
        & 51.9$\pm$2.6 & 54.3$\pm$3.1  & 53.0$\pm$3.8        \\ \hline
       \multirow{2}{*}{AE} & 98.8$\pm$0.8 & 98.5$\pm$0.8 &  98.2$\pm$1.0     \\ & 54.8$\pm$2.3 & 54.8$\pm$2.4  & 55.7$\pm$3.3        \\ \hline
       
       \multirow{2}{*}{AM}  & 98.7$\pm$0.9 &  97.8$\pm$1.0 & 96.9$\pm$1.4       \\ 
       & 51.7$\pm$2.9 & 54.4$\pm$5.5  & 55.9$\pm$3.9        \\ \hline
    \end{tabular}
    \caption{\textit{Generalization onto unknown} metric (top) and \textit{identifiability} (bottom).
    }
    \label{tab:unknown}
\end{table}

\section{Discussion}

By proposing two different metrics, we believe we quantify two different behaviors of a couple model - dataset: how nuisance information entangled in the data  is used to infer (\textit{GTB}) and how much generic information is learned by the model from the data (\textit{GU}). 

\textit{GTB} seems to have a more desirable behavior since it has the same monotony as the mutual information and $\beta$. It reflects the intuition that the problem becomes harder while dependence of label and features on the nuisance factor increases. \textit{GU} metric also decreases as the mutual information increases. Besides, \textit{GU} metric provides a good \textit{a priori} estimation of model performance on data  relying in a new domain (i.e. a new nuisance factor value), unlike \textit{identifiability}.

From \textit{GTB} experiment, it appears clearly that representation invariance, estimated with \textit{identifiability}, does not guarantee model invariance. We are aware that we have investigated a particular case, but we still express our concerns on adversarial methods which enforce, by distribution matching, representation to remain invariant. In this context, the invariance loss \cite{ganin2014unsupervised,xie2017controllable} seems intractable and not a good proxy of model invariance. It is mainly due to large number of values the nuisance factor takes as underlined in \cite{moyer2018evading, feutry2018learning}.

In a future work, we want to explore the decomposition $\mathbb P(X,Y) = \mathbb E_{S}[\mathbb P(X|S) \mathbb P(Y|X,S)]$ in order to build a third invariance metric based on the modification of $\mathbb P(X|S)$. We did not succeed yet in formulating relevant train / test split for quantifying this change in the data. We believe the data filtering approach we have suggested is a first proxy of it.
\bibliography{naaclhlt2019}

\begin{thebibliography}{16}
\expandafter\ifx\csname natexlab\endcsname\relax\def\natexlab#1{#1}\fi

\bibitem[{Bengio et~al.(2006)Bengio, Delalleau, and Roux}]{bengio2006curse}
Yoshua Bengio, Olivier Delalleau, and Nicolas~L Roux. 2006.
\newblock The curse of highly variable functions for local kernel machines.
\newblock In \emph{Advances in neural information processing systems}, pages
  107--114.

\bibitem[{Bengio et~al.(2009)}]{bengio2009learning}
Yoshua Bengio et~al. 2009.
\newblock Learning deep architectures for ai.
\newblock \emph{Foundations and trends{\textregistered} in Machine Learning},
  2(1):1--127.

\bibitem[{Blitzer et~al.(2011)Blitzer, Kakade, and Foster}]{blitzer2011domain}
John Blitzer, Sham Kakade, and Dean Foster. 2011.
\newblock Domain adaptation with coupled subspaces.
\newblock In \emph{Proceedings of the Fourteenth International Conference on
  Artificial Intelligence and Statistics}, pages 173--181.

\bibitem[{Dheeru and Karra~Taniskidou(2017)}]{Dua:2017}
Dua Dheeru and Efi Karra~Taniskidou. 2017.
\newblock \href {http://archive.ics.uci.edu/ml} {{UCI} machine learning
  repository}.

\bibitem[{Feutry et~al.(2018)Feutry, Piantanida, Bengio, and
  Duhamel}]{feutry2018learning}
Cl{\'e}ment Feutry, Pablo Piantanida, Yoshua Bengio, and Pierre Duhamel. 2018.
\newblock Learning anonymized representations with adversarial neural networks.
\newblock \emph{arXiv preprint arXiv:1802.09386}.

\bibitem[{Glorot et~al.(2011)Glorot, Bordes, and Bengio}]{glorot2011domain}
Xavier Glorot, Antoine Bordes, and Yoshua Bengio. 2011.
\newblock Domain adaptation for large-scale sentiment classification: A deep
  learning approach.
\newblock In \emph{Proceedings of the 28th international conference on machine
  learning (ICML-11)}, pages 513--520.

\bibitem[{Joulin et~al.(2016)Joulin, Grave, Bojanowski, and
  Mikolov}]{joulin2016bag}
Armand Joulin, Edouard Grave, Piotr Bojanowski, and Tomas Mikolov. 2016.
\newblock Bag of tricks for efficient text classification.
\newblock \emph{arXiv preprint arXiv:1607.01759}.

\bibitem[{Kamiran and Calders(2009)}]{kamiran2009classifying}
Faisal Kamiran and Toon Calders. 2009.
\newblock Classifying without discriminating.
\newblock In \emph{Computer, Control and Communication, 2009. IC4 2009. 2nd
  International Conference on}, pages 1--6. IEEE.

\bibitem[{Kamishima et~al.(2011)Kamishima, Akaho, and
  Sakuma}]{kamishima2011fairness}
Toshihiro Kamishima, Shotaro Akaho, and Jun Sakuma. 2011.
\newblock Fairness-aware learning through regularization approach.
\newblock In \emph{Data Mining Workshops (ICDMW), 2011 IEEE 11th International
  Conference on}, pages 643--650. IEEE.

\bibitem[{Krizhevsky et~al.(2012)Krizhevsky, Sutskever, and
  Hinton}]{krizhevsky2012imagenet}
Alex Krizhevsky, Ilya Sutskever, and Geoffrey~E Hinton. 2012.
\newblock Imagenet classification with deep convolutional neural networks.
\newblock In \emph{Advances in neural information processing systems}, pages
  1097--1105.

\bibitem[{McAuley and Leskovec(2013)}]{mcauley2013hidden}
Julian McAuley and Jure Leskovec. 2013.
\newblock Hidden factors and hidden topics: understanding rating dimensions
  with review text.
\newblock In \emph{Proceedings of the 7th ACM conference on Recommender
  systems}, pages 165--172. ACM.

\bibitem[{Moyer et~al.(2018)Moyer, Gao, Brekelmans, Steeg, and
  Galstyan}]{moyer2018evading}
Daniel Moyer, Shuyang Gao, Rob Brekelmans, Greg~Ver Steeg, and Aram Galstyan.
  2018.
\newblock Evading the adversary in invariant representation.
\newblock \emph{arXiv preprint arXiv:1805.09458}.

\bibitem[{Pan and Yang(2010)}]{pan2010survey}
Sinno~Jialin Pan and Qiang Yang. 2010.
\newblock A survey on transfer learning.
\newblock \emph{IEEE Transactions on knowledge and data engineering},
  22(10):1345--1359.

\bibitem[{Quionero-Candela et~al.(2009)Quionero-Candela, Sugiyama,
  Schwaighofer, and Lawrence}]{quionero2009dataset}
Joaquin Quionero-Candela, Masashi Sugiyama, Anton Schwaighofer, and Neil~D
  Lawrence. 2009.
\newblock \emph{Dataset shift in machine learning}.
\newblock The MIT Press.

\bibitem[{Xie et~al.(2017)Xie, Dai, Du, Hovy, and Neubig}]{xie2017controllable}
Qizhe Xie, Zihang Dai, Yulun Du, Eduard Hovy, and Graham Neubig. 2017.
\newblock Controllable invariance through adversarial feature learning.
\newblock In \emph{Advances in Neural Information Processing Systems}, pages
  585--596.

\bibitem[{Zemel et~al.(2013)Zemel, Wu, Swersky, Pitassi, and
  Dwork}]{zemel2013learning}
Rich Zemel, Yu~Wu, Kevin Swersky, Toni Pitassi, and Cynthia Dwork. 2013.
\newblock Learning fair representations.
\newblock In \emph{International Conference on Machine Learning}, pages
  325--333.

\end{thebibliography}
\bibliographystyle{acl_natbib}

\newpage {}\newpage
\appendix

\section{Data filtering details}
\label{sec:data_filtering}
We provide details about our data filtering approach. First, the contribution of mutual information of $i(X,S)$ is defined as follows:
\begin{equation*}
    I_{\mathbb P}(X,S) = \mathbb E_X[i(X)]]
\end{equation*}
where
\begin{equation*}
    i(X) = \mathbb E_S[- \log d \mathbb P(S)] -\mathbb E_{S|X} [- \log d \mathbb P(S|X)]
\end{equation*}
We used an approximate lower bound of the mutual information $I_{\mathbb P}(X,S) = H_{\mathbb P}(S) -H_{\mathbb P}(S|X)$. For a given estimation $\hat P (S|X)$ of $\mathbb P(S|X)$, we can lower bound: 

$$H_{\mathbb P}(S|X) \leq \mathbb E_{(X,S)\sim \mathbb P} [- \log d\hat{\mathbb P}(S|X)] $$
since the KL divergence is positive and approximate: 
$$H_{\mathbb P}(S) \approx H_{\hat{\mathbb P}}(S) $$
where $\hat{\mathbb P}(S)$ is a count measure. Then: 
$$I_{\mathbb P}(X,S) \gtrapprox H_{\hat{\mathbb P}}(S) - \mathbb E_{(X,S)\sim \mathbb P} [- \log d\hat{\mathbb P}(S|X)]$$
The estimation of $\mathbb P(S|X)$ was done using a simple text classification model \cite{joulin2016bag} with the same setup described in \ref{sec:training_details}.

\begin{algorithm}[h!]
\caption{Generalization under target bias split with respect to $\mathbb P_\beta $}\label{alg:target_bias}
\begin{algorithmic}[1]
\State \textbf{input} $\mathcal D = (x_n,y_n, s_n)_{i=1}^N$, $\mathcal Y$, $\mathcal S$, $\beta$
\State Split $\mathcal D$ into $\mathcal D_{\mathrm{tr}}$ and $\mathcal D_{\mathrm{ts}}$
\State $\mathcal D \leftarrow \mathcal D_{\mathrm{tr}}$
\State $\mathcal D_\mathrm{tr} \leftarrow \{\}$
\While{$|\mathcal S| > 0$}
\State $\mathcal S' \leftarrow $ pick at random $(s_1, ..., s_{|\mathcal Y|}) \subset \mathcal S$ 
\State $N \leftarrow \min(|\mathcal D_{y,s}|, y \in \mathcal Y, s \in \mathcal S') $
\State $\mathcal Y' \leftarrow \mathcal Y$
\For{$s \in \mathcal S'$}
\State $y \leftarrow$ pick at random in $\mathcal Y'$
\State $\mathcal D' \leftarrow$ pick at random $\mathrm{int}(\beta N)$ 
\State ~~~~~~~~~~samples from $\mathcal D_{y,s}$
\State $\mathcal D_\mathrm{tr} \leftarrow \mathcal  D_\mathrm{tr} \cup \mathcal D'$
\State $\mathcal Y' \leftarrow \mathcal Y' \backslash \{y\}$
\For{$y \in \mathcal Y'$}
\State $\mathcal D' \leftarrow $ pick at random $\mathrm{int}(\frac{1-\beta}{|\mathcal Y| -1 }N)$ \State ~~~~~~~~~~samples from $\mathcal D_{y,s}$
\State $\mathcal D_\mathrm{tr} \leftarrow  \mathcal D_\mathrm{tr} \cup \mathcal D'$
\EndFor 
\EndFor
\State $\mathcal S \leftarrow \mathcal S \backslash \mathcal S'$
\EndWhile
\State \textbf{return} $(\mathcal D_{\mathrm tr}, \mathcal D_\mathrm{ts})$
\end{algorithmic}
\end{algorithm}

\begin{algorithm}[h!]
\caption{Rejection}\label{alg:rejection}
\begin{algorithmic}[1]
\State \textbf{input} $\mathcal D = (x_n,y_n, s_n)_{1 \leq i \leq N}$, $N_\mathrm{it}$, $N_b$, $\alpha$, $\mathcal Y$, $\mathcal S$
\State $\mathrm{Seq} \leftarrow \{ \mathcal D\}$
\For{$N_\mathrm{it}$ iterations}
\State Split $\mathcal D$ into $N_b$ subsets $(\mathcal D_{i})_{i=1}^{N_b}$
\State $\mathcal I \leftarrow \{\}$
\For{$1 \leq  n  \leq N_b$}
\State Train $\hat{\mathbb P}(S)$ and $\hat{\mathbb P}(S|X)$ on $\cup_{i \neq n} \mathcal D_i$
\State $\mathcal I \leftarrow \mathcal I \cup \{i(x)\backslash (x,y,s) \in \mathcal D_n\}$
\EndFor
\State $\varepsilon \leftarrow \mathrm{quantile}(\mathcal I, \alpha)$
\State $\mathcal R \leftarrow \{\}$
\For{$s \in \mathcal S$}
\State $n_s \leftarrow \min (|\{(x,y,s) \in \mathcal D_{y,s} $ 
\State $ ~~~~~~~~~~\backslash i_{\hat{\mathbb P}}(x) < \varepsilon\} |, y \in \mathcal Y)$
\For{$y \in \mathcal Y$}
\State $\mathcal D_{s,y} \leftarrow \mathrm{sort}(\mathcal D_{s,y})$ by
\State ~~~~~~ascending value of $i_{\hat{\mathbb P}}(x)$
\State $\mathcal R \leftarrow \mathcal R \cup \mathcal D_{s,y}[:n_s]$
\EndFor
\EndFor
\State $\mathcal D \leftarrow \mathcal D \backslash \mathcal R$
\State $\mathrm{Seq} \leftarrow \mathrm{Seq} \cup \{ \mathcal D\}$
\EndFor
\State \textbf{return} $\mathrm{Seq}$
\end{algorithmic}
\end{algorithm}
In algorithm \ref{alg:rejection}, $\mathcal D_{y,s}$ for $(s,y) \in \mathcal S \times \mathcal Y$ as follows:
$$\mathcal D_{y,s} = \{(x',y',s') \in \mathcal D \backslash y'=y, s'=s\}$$

\section{Training details}
\label{sec:training_details}
\paragraph{Text classification baseline studied} We used recommended hyper-parameters of the original paper: 20 dimensions of word embedding including bi-grams trained during 5 epochs with a learning rate of $0.25$.
\paragraph{Compute \textit{identifiability}} We used a simple Multi-Layer-Perceptron on top of considered representation. The two first layers are 100 then 200 neurons with ReLU activation and the last layer is a softmax layers on the sources for \textit{GBT} and old VS new value of the nuisance factor for \textit{GU}. The net is trained during 10 epochs with an Adam optimizer minimizing categorical cross-entropy with batch of size 128.

\newpage {} \newpage
\section{Additional figures}

\begin{figure*}
\begin{center}
\footnotesize

\begin{tikzpicture}
	\begin{axis}[
		height=4.cm,
		width=8cm,
		grid=major,
        legend pos=south west
	]
	\addplot coordinates {
	    (0.6, 100)
		(0.7, 98.2)
        (0.8, 96.2)
        (0.9, 91.2)
        (1.0, 86.2)};
	 \label{0}
	
	\addplot coordinates {
	    (0.6, 98.9)
		(0.7, 96.7)
        (0.8, 94.2)
        (0.9, 90.7)
        (1.0, 85.7)};
    \label{1}
	
	\addplot coordinates {
	    (0.6, 100)
		(0.7, 96.1)
        (0.8, 91.4)
        (0.9, 83.5)
        (1.0, 74)};
	\label{2}
	
	\label{3}
	\end{axis}
\end{tikzpicture}
\begin{tikzpicture}
	\begin{axis}[
		height=4.cm,
		width=5.25cm,
		grid=major,
        legend pos=south west
	]
	\addplot coordinates {
	    (0.6, 76.5)
		(0.7, 73.2)
        (0.8, 72)
        (0.9, 72.4)
        (1.0, 69)};
	 \label{0}
	
	\addplot coordinates {
	    (0.6, 76.5)
		(0.7, 78.25)
        (0.8, 80.8)
        (0.9, 74.3)
        (1.0, 70.3)};
    \label{1}
	
	\addplot coordinates {
	    (0.6, 95)
		(0.7, 96)
        (0.8, 92.3)
        (0.9, 90)
        (1.0, 82.3)};
	\label{2}

	\end{axis}
\end{tikzpicture}
\begin{tikzpicture}
	\begin{axis}[
		height=4.cm,
		width=5.25cm,
		grid=major,
        legend pos=south west
	]
	\addplot coordinates {
	    (0.6, 15.3)
		(0.7, 29.3)
        (0.8, 43.2)
        (0.9, 57.9)
        (1.0, 69.0)};
	 \label{0}
	
	\addplot coordinates {
	    (0.6, 15.3)
		(0.7, 31.3)
        (0.8, 48.5)
        (0.9, 59.4)
        (1.0, 70.3)};
    \label{1}
	
	\addplot coordinates {
	    (0.6, 19.0)
		(0.7, 38.4)
        (0.8, 55.4)
        (0.9, 72.0)
        (1.0, 82.3)};
	\label{2}
	\end{axis}
\end{tikzpicture}

\begin{tikzpicture}
	\begin{axis}[
		height=4.cm,
		width=8cm,
		grid=major,
        legend pos=south west
	]
	\addplot coordinates {
	    (0.6, 99.7)
		(0.7, 98.1)
        (0.8, 96.)
        (0.9, 91.8)
        (1.0, 85)};
	 \label{0}
	
	\addplot coordinates {
	    (0.6, 99.2)
		(0.7, 97.1)
        (0.8, 93.8)
        (0.9, 88.6)
        (1.0, 78.0)};
    \label{1}
	
	\addplot coordinates {
	    (0.6, 99.4)
		(0.7, 96.9)
        (0.8, 93.2)
        (0.9, 86.3)
        (1.0, 71.1)};
	\label{2}
	
	\label{3}
	\end{axis}
\end{tikzpicture}
\begin{tikzpicture}
	\begin{axis}[
		height=4.cm,
		width=5.25cm,
		grid=major,
        legend pos=south west
	]
	\addplot coordinates {
	    (0.6, 84)
		(0.7, 84.8)
        (0.8, 87.2)
        (0.9, 85.6)
        (1.0, 81.7)};
	 \label{0}
	
	\addplot coordinates {
	    (0.6, 99.5)
		(0.7, 94.5)
        (0.8, 91.3)
        (0.9, 90.25)
        (1.0, 85.1)};
    \label{1}
	
	\addplot coordinates {
	    (0.6, 101)
		(0.7, 97.25)
        (0.8, 100)
        (0.9, 92.125)
        (1.0, 88.4)};
	\label{2}

	\end{axis}
\end{tikzpicture}
\begin{tikzpicture}
	\begin{axis}[
		height=4.cm,
		width=5.25cm,
		grid=major,
        legend pos=south west
	]
	\addplot coordinates {
	    (0.6, 16.8)
		(0.7, 35.1)
        (0.8, 52.3)
        (0.9, 68.5)
        (1.0, 81.7)};
	 \label{0}
	
	\addplot coordinates {
	    (0.6, 19.9)
		(0.7, 37.8)
        (0.8, 54.8)
        (0.9, 72.2)
        (1.0, 85.1)};
    \label{1}
	
	\addplot coordinates {
	    (0.6, 20.2)
		(0.7, 38.9)
        (0.8, 60)
        (0.9, 73.7)
        (1.0, 88.4)};
	\label{2}
	\end{axis}
\end{tikzpicture}

\begin{tikzpicture}
	\begin{axis}[
		height=4.cm,
		width=8cm,
		grid=major,
        legend pos=south west
	]
	\addplot coordinates {
	    (0.6, 99.3)
		(0.7, 97.4)
        (0.8, 95.3)
        (0.9, 91.3)
        (1.0, 83.3)};
	 \label{0}
	
	\addplot coordinates {
	    (0.6, 99.)
		(0.7, 97.3)
        (0.8, 94.7)
        (0.9, 87.5)
        (1.0, 76.5)};
    \label{1}
	
	\addplot coordinates {
	    (0.6, 99.5)
		(0.7, 98.8)
        (0.8, 91.5)
        (0.9, 82.7)
        (1.0, 68.1)};
	\label{2}
	\end{axis}
\end{tikzpicture}
\begin{tikzpicture}
	\begin{axis}[
		height=4.cm,
		width=5.25cm,
		grid=major,
        legend pos=south west
	]
	\addplot coordinates {
	    (0.6, 90.5)
		(0.7, 92.5)
        (0.8, 87.5)
        (0.9, 86.8)
        (1.0, 85.1)};
	 \label{0}
	
	\addplot coordinates {
	    (0.6, 91.5)
		(0.7, 96)
        (0.8, 92.5)
        (0.9, 92.1)
        (1.0, 88)};
    \label{1}
	
	\addplot coordinates {
	    (0.6, 103)
		(0.7, 104.8)
        (0.8, 104.2)
        (0.9, 99.9)
        (1.0, 92.1)};
	\label{2}

	\end{axis}
\end{tikzpicture}
\begin{tikzpicture}
	\begin{axis}[
		height=4.cm,
		width=5.25cm,
		grid=major,
        legend pos=south west
	]
	\addplot coordinates {
	    (0.6, 18.1)
		(0.7, 37)
        (0.8, 52.5)
        (0.9, 69.4)
        (1.0, 85.1)};
	 \label{0}
	
	\addplot coordinates {
	    (0.6, 18.3)
		(0.7, 38.4)
        (0.8, 55.5)
        (0.9, 73.7)
        (1.0, 88)};
    \label{1}
	
	\addplot coordinates {
	    (0.6, 20.6)
		(0.7, 41.9)
        (0.8, 62.5)
        (0.9, 79.9)
        (1.0, 92.1)};
	\label{2}
	\end{axis}
\end{tikzpicture}

\begin{tikzpicture}
	\begin{axis}[
		height=4.cm,
		width=8cm,
		grid=major,
        legend pos=south west
	]
	\addplot coordinates {
	    (0.6, 99.7)
		(0.7, 99)
        (0.8, 96.7)
        (0.9, 94.2)
        (1.0, 88.3)};
	 \label{0}
	
	\addplot coordinates {
	    (0.6, 99.8)
		(0.7, 98.7)
        (0.8, 96.3)
        (0.9, 93.1)
        (1.0, 86.3)};
    \label{1}
	
	\addplot coordinates {
	    (0.6, 95.2)
		(0.7, 98.4)
        (0.8, 94.7)
        (0.9, 89.8)
        (1.0, 76.9)};
	\label{2}
	\end{axis}
\end{tikzpicture}
\begin{tikzpicture}
	\begin{axis}[
		height=4.cm,
		width=5.25cm,
		grid=major,
        legend pos=south west
	]
	\addplot coordinates {
	    (0.6, 87.5)
		(0.7, 93)
        (0.8, 90.2)
        (0.9, 89.125)
        (1.0, 86.3)};
	 \label{0}
	
	\addplot coordinates {
	    (0.6, 92.5)
		(0.7, 96.75)
        (0.8, 93)
        (0.9, 89.9)
        (1.0, 86.7)};
    \label{1}
	
	\addplot coordinates {
	    (0.6, 101.5)
		(0.7, 97.25)
        (0.8, 95)
        (0.9, 94.5)
        (1.0, 90.3)};
	\label{2}

	\end{axis}
\end{tikzpicture}
\begin{tikzpicture}
	\begin{axis}[
		height=4.cm,
		width=5.25cm,
		grid=major,
        legend pos=south west
	]
	\addplot coordinates {
	    (0.6, 17.5)
		(0.7, 37.2)
        (0.8, 54.1)
        (0.9, 71.3)
        (1.0, 86.3)};
	 \label{0}
	
	\addplot coordinates {
	    (0.6, 18.5)
		(0.7, 38.7)
        (0.8, 55.8)
        (0.9, 71.9)
        (1.0, 86.7)};
    \label{1}
	
	\addplot coordinates {
	    (0.6, 20.3)
		(0.7, 38.9)
        (0.8, 57)
        (0.9, 75.6)
        (1.0, 90.3)};
	\label{2}
	\end{axis}
\end{tikzpicture}

\end{center}
\caption{From left to right: \textit{Generalization under target bias}, $\mathrm{disc}(y)/(2\beta-1)$ and $\mathrm{disc(1)}$ according to $\beta$ (iterations: \{$\mathcal D_0$: \ref{0}, $\mathcal D_1$: \ref{1}, $\mathcal D_2$: \ref{2}\}). From the top to the bottom: Yelp, AB, AM, AE.}
\label{fig:target_bias_appendix}
\end{figure*}
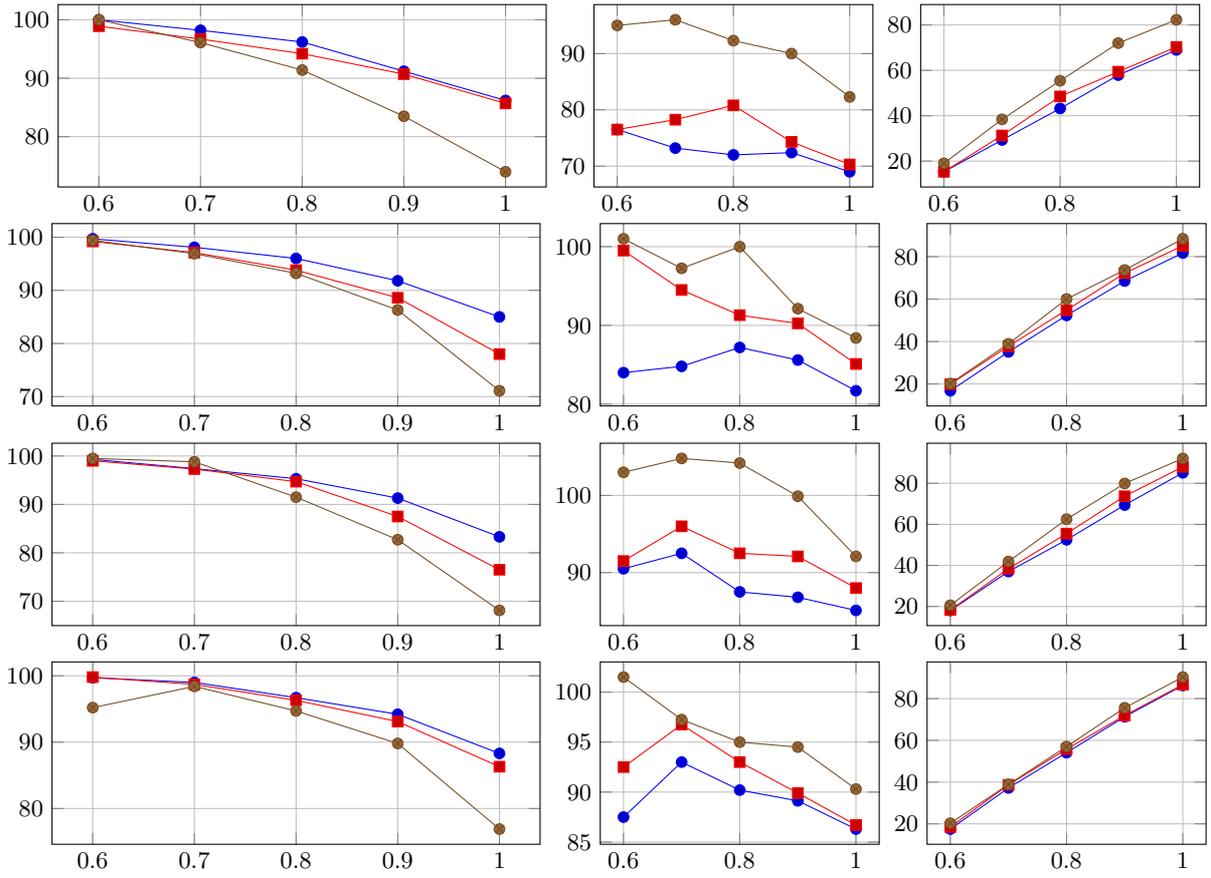

\end{document}